\documentclass[letterpaper, 10 pt, journal, twoside]{IEEEtran/ieeetran}
\usepackage{graphics} 
\usepackage{graphics} 

\usepackage{times} 
\usepackage{amsmath} 
\usepackage{amssymb}  
\usepackage{dsfont}  
\usepackage{mathtools}
\usepackage{hyperref}
\usepackage[normalem]{ulem}
\useunder{\uline}{\ul}{}
\usepackage{multirow}
\usepackage{hyperref}

\usepackage{graphicx} 
\usepackage{subcaption}

\usepackage{pifont}
\newcommand{\cmark}{\ding{51}}%

\usepackage[dvipsnames]{xcolor}
\usepackage{xspace}

\newcommand{\myparagraph}[1]{\vspace{3pt}\noindent\textbf{#1}}

\DeclareMathOperator*{\argmin}{arg\,min}

\IEEEoverridecommandlockouts           

\title{Boosting Deep Open World Recognition\\ by Clustering} 

\author{Dario Fontanel$^{*,1}$, Fabio Cermelli$^{*,1,2}$, Massimiliano Mancini$^{3}$,\\ \hspace{2em}Samuel Rota Bulò$^{4}$, Elisa Ricci$^{5}$ and Barbara Caputo$^{1,2}$
\thanks{Manuscript received: February, 24, 2020; Revised  April, 22, 2020; Accepted June, 19, 2020.}
\thanks{This paper was recommended for publication by Editor Cesar Cadena Lerma upon evaluation of the Associate Editor and Reviewers' comments.}
\thanks{} 
\thanks{*indicates equal contribution}
\thanks{$^{1}$D. Fontanel, F. Cermelli and B. Caputo are with Politecnico di Torino, Turin, Italy. {\tt\small \{dario.fontanel, fabio.cermelli, barbara.caputo\}@polito.it}}%
\thanks{$^{2}$F. Cermelli and B. Caputo are with Italian Institute of Technology, Genoa, Italy.}
\thanks{$^{3}$M. Mancini is with Sapienza University of Rome, Rome, Italy and University of T\"ubingen, T\"ubingen, Germany. {\tt\small mancini@diag.uniroma1.it}}
\thanks{$^{4}$S. Rota Bulò is with Mapillary, Graz, Austria. {\tt\small samuel@mapillary.com}}
\thanks{$^{5}$E. Ricci is with Fondazione Bruno Kessler, Trento, Italy and  University of Trento, Trento, Italy. {\tt\small eliricci@fbk.eu}}
\thanks{Digital Object Identifier (DOI): see top of this page.}
}

\begin{document}

\maketitle

\begin{abstract}

While convolutional neural networks have brought significant advances in robot vision, their ability is often limited to  
\textit{closed world} scenarios, where the number of semantic concepts to be recognized is determined by the available training set. Since it is practically impossible to capture all possible semantic concepts present in the real world in a single training set, we need to break the closed world assumption, equipping our robot with the capability to act in an \textit{open world}. To provide such ability, a robot vision system should be able to (i) identify whether an instance does not belong to the set of known categories (i.e. open set recognition), and (ii) extend its knowledge to learn new classes over time (i.e. incremental learning).
In this work, we show how we can boost the performance of deep open world recognition algorithms by means of a new loss formulation enforcing a global to local clustering of class-specific features. In particular, a first loss term, i.e. \textit{global clustering}, forces the network to map samples closer to the class centroid they belong to 
while the second one, \textit{local clustering}, shapes the representation space in such a way that samples of the same class get closer in the representation space while pushing away neighbours belonging to other classes.
Moreover, we propose a strategy to learn class-specific rejection thresholds, instead of heuristically estimating a single global threshold, as in previous works. 
Experiments on three benchmarks 
show the effectiveness of our approach. 

\end{abstract}


\section{Introduction} 

\IEEEPARstart{A}{}long-standing goal of artificial intelligence and robotics is implementing agents able to interact in the real world. In order to achieve this goal, a crucial step is making the agent able to understand the current state of the surrounding environment. Within this context, visual cameras are one of the most powerful and information-rich sensors, thus a lot of research efforts have been spent on improving robot vision systems. Due to their effectiveness in addressing visual problems, deep neural networks have been used in many robotic tasks such as egomotion estimation \cite{costante2016exploring}, depth prediction \cite{jafari2017analyzing,mancini2017toward}, object grasping \cite{johns2016deep, levine2018learning} and semantic segmentation \cite{schwarz2018rgb,oliveira2018efficient}. 
Despite their effectiveness, deep neural networks limit their understanding to the particular set of knowledge present in the training set they are tuned on, relying on the \textit{closed world assumption}  (CWA). Obviously, this is a fundamental drawback if we want to apply any visual system, especially a recognition based one, in the real world. Indeed, the world contains an infinite set of possible input conditions (e.g. various illumination, environments) and semantic concepts: capturing them in a single training set is practically unfeasible. Under these perspectives, we would like to make our algorithm both robust to unseen input conditions as well as being able to detect and learn novel semantic concepts. While previous work tried to address the first problem in the context of domain adaptation \cite{wulfmeier2017addressing,wulfmeier2018incremental,mancini2018kitting} and generalization \cite{mancini2018robust}, little attention has been posed to the second one. Here we show how we can break the CWA developing a visual system able to work in the \textit{open world}.

\begin{figure}[tb]
  \centering
  \includegraphics[width=\columnwidth]{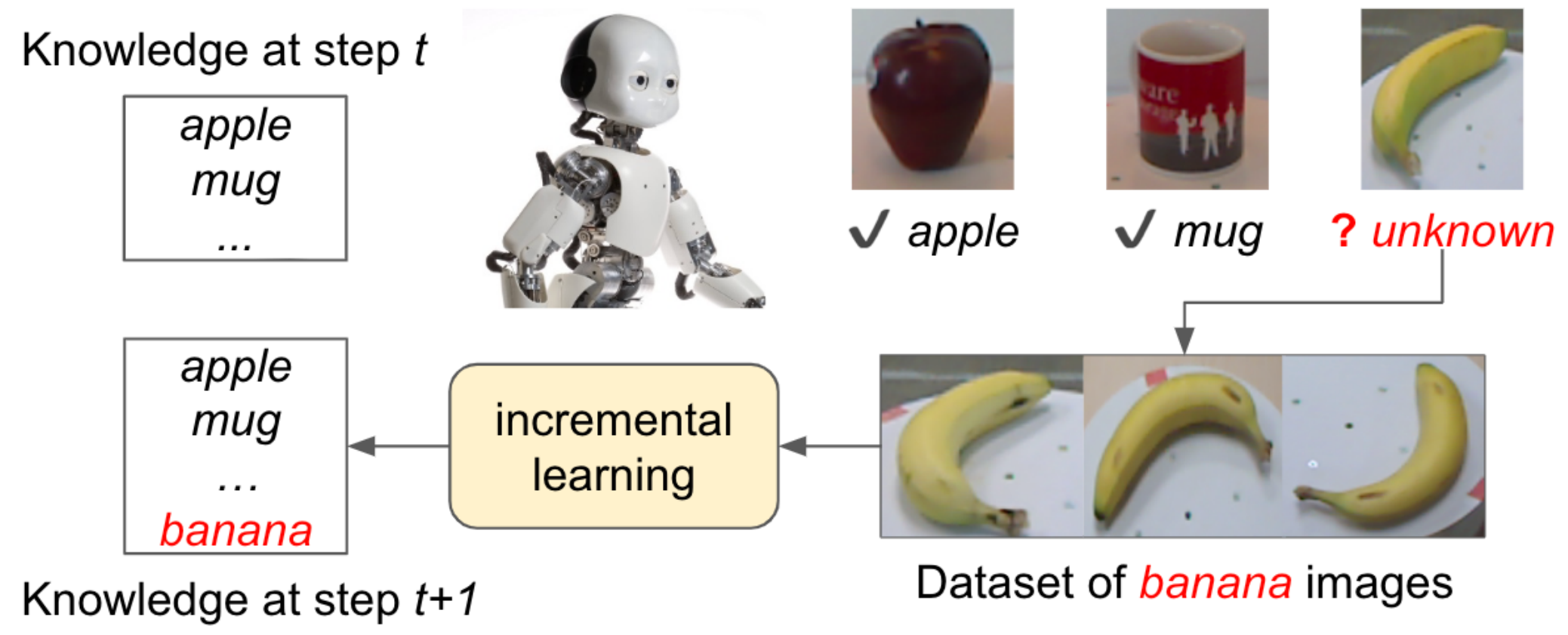} 
  \vspace{-15pt}
   \caption{In the open world scenario a robot must be able to classify correctly known objects, (\textit{apple} and \textit{mug}), and detect novel semantic concepts (e.g. \textit{banana}). When a novel concept is detected, it should learn the new class from an auxiliary dataset, updating its internal knowledge.}
  \vspace{-20pt}
  \label{fig:teaser}
 \end{figure}
 
To clarify our goal, let us consider the example shown in Fig.~\ref{fig:teaser}. The robot has a knowledge base composed by a limited number of classes. Given an image containing an unknown concept (e.g. banana), we want the robot to detect it as unknown and being able to add it to its knowledge base in subsequent learning stages. 
To accomplish this goal, it is very important for a robot vision system to have two crucial abilities: (i) it must be able to recognize already seen concepts and detect unknown ones 
(i.e. open set recognition), and (ii) it must be able to extend its knowledge base with new classes 
(i.e. incremental learning), without forgetting the already learned ones and without access to old training sets, avoiding \textit{catastrophic forgetting} \cite{mccloskey1989catastrophic}). 
While open set recognition \cite{scheirer2012toward, fragoso2013evsac, li2005open} and incremental learning \cite{rebuffi2017icarl,camoriano2016incremental,camoriano2017incremental,valipour2017incremental} are well-studied problems in the literature, few works proposed a solution to solve them together \cite{bendale2015towards,de2016online,mancini2019knowledge}.  Standard approaches for open world recognition (OWR) equip the nearest class mean (NCM) classification algorithm with a rejection option based on an estimated threshold. While standard approaches \cite{bendale2015towards,de2016online} use shallow features, only recently it has been showed how deep neural networks can be successfully employed also in the OWR scenario \cite{mancini2019knowledge}. In this work we follow the deep learning based approach of \cite{mancini2019knowledge} but we take a step forward. 
In fact, we argue that it is crucial to force the deep architecture used as feature extractor 
to cluster appropriately samples belonging to the same class, while pushing away samples of other classes.
For this reason, we introduce a global clustering loss term that aims at keeping closer the features of samples belonging to the same class to their class centroid. Furthermore, we show how the soft nearest neighbor loss \cite{salakhutdinov2007learning, frosst2019analyzing} can be successfully employed as a local clustering loss term in order to force pair of samples of the same class to be closer in the learned metric space than points of other classes. 
Additionally, differently from previous works \cite{bendale2015towards, mancini2019knowledge} 
we avoid to estimate a global rejection threshold on the model predictions based on heuristic rules but we (i) define an independent threshold for each class and 
(ii) we explicitly learn the thresholds by using a margin-based loss function which balances rejection errors on samples of a reserved memory held-out from the training. 
We evaluate the effectiveness of our method on Core50 \cite{lomonaco2017core50}, RGB-D Object Dataset \cite{lai2011large} and CIFAR-100 \cite{krizhevsky2009learning} datasets, showing that introducing the two complementary clustering loss and learning the rejection thresholds outperforms previous approaches.

\myparagraph{Contributions.} To summarize, the contributions of this paper are as follows:
\begin{itemize}
    \item {We introduce two clustering losses to 
    effectively localize samples of the same class in the representation space, while separating them from points belonging to other classes;}
    \item We propose an effective method to detect unknown samples based on learned class-specific rejection thresholds; 
    \item We demonstrate the superiority of our method over state of the art, {reporting a quantitative analysis and an extensive ablation of the components of our model.}
\end{itemize}

\section{Related Works} 

The necessity of breaking the CWA for robot vision systems \cite{sunderhauf2018limits} has lead various research efforts on understanding how to extend pre-trained models with new semantic concepts while retain previous knowledge. To this extent, recent years have seen a growing interest on topics such as continual \cite{lesort2020continual} and incremental learning \cite{valipour2017incremental,camoriano2017incremental,cermelli2019rgb}. In \cite{pasquale2015teaching}, the authors study how to update the visual recognition system of a humanoid robot on multiple training sessions. 
In \cite{camoriano2017incremental}, a variant of the Regularized Least Squares algorithm is introduced to add new classes to a pre-trained model. 
In \cite{parisi2018lifelong}, a growing dual-memory is proposed to dynamically learn novel object instances and categories. 
In \cite{lagunes2019learning} the authors proposed to learn an embedding in order to perform fast incremental learning of new objects. Another solution to this problem can exploit the help of a human-robot interaction, as in \cite{valipour2017incremental} where a robot incrementally learns to detect new objects as they are manually pointed by a human. 

While these approaches focus on incremental and continual learning, acting in the open world requires both detecting unknown concepts automatically and adding them in subsequent learning stages. 
Towards this objective, in \cite{bendale2015towards} the authors introduced the OWR setting, as a more general and realistic scenario for agents acting in the real world. 
In \cite{bendale2015towards}, the authors extend the Nearest Class Mean (NCM) classifier \cite{mensink2012metric,guerriero2018deep}
 to act in the open set scenario, proposing the Nearest Non-Outlier algorithm (NNO). In order to estimate whereas a test sample belongs to the known or unknown set of categories, this method introduces a {rejection} threshold that, after the first initialization phase, is kept fixed for subsequent learning episodes. In \cite{de2016online}, the authors proposed to tackle OWR with the Nearest Ball Classifier, with a rejection threshold based on the confidence of the predictions. Recently, in \cite{mancini2019knowledge}, the NNO algorithm of \cite{bendale2015towards} has been extended by employing an end-to-end trainable deep architecture as feature extractor, with a dynamic update strategy for the rejection threshold.
In this work, we show how we can improve the performances of NCM based classifier for OWR through a global to local clustering loss. Moreover, differently for previous works, our rejection threshold is class-specific and is explicitly learned rather than fixed based on heuristic strategies. 

\section{Our Method}
In this section we describe our OWR method. We start by formalizing the OWR problem and describing the DeepNNO framework \cite{mancini2019knowledge} which serves as our starting point. We then discuss our core components, the global to local clustering and how we learn the class-specific rejection thresholds. 

\subsection{Problem Definition}
The goal of OWR is producing a model capable of (i) recognizing known concepts (i.e. classes seen during training), (ii) detecting unseen categories (i.e. classes not present in any training set used for training the model) and (iii) incrementally add new classes as new training data is available. 
Formally, let us denote as $\mathcal{X}$ and $\mathcal{K}$ the input space (i.e. image space) and the closed world output space respectively (i.e. set of known classes).  Moreover, since our output space will change as we receive new data containing novel concepts, we will denote as $\mathcal{K}_t$ the set of classes seen after the $t_{\text{th}}$ incremental step, with $\mathcal{K}_0$ denoting the category present in the first training set. Additionally, since we aim to detect if an image contains an unknown concept, in the following we will denote as $unk$ the special unknown class, building the output space as $\mathcal{K}_t \cup \{unk\}$. We assume that, at each incremental step, we have access to a training set $\mathcal{T}_t = \{(x^t_1 , c^t_1), \cdots, (x^t_{N_t}, c^t_{N_t}) \}$, with $N_t=|\mathcal{T}_t|$, $x^t \in \mathcal{X}$, and $c^t \in \mathcal{C}_t$, where $\mathcal{C}_t$ is the set of categories contained in the training set $\mathcal{T}_t$.
Note that, without loss of generality, in each incremental step, we assume to see a new set of classes $\mathcal{C}_i\cap\mathcal{C}_j = \emptyset$ if $i\neq j$. The set of known classes at step $t$ is computed as $\mathcal{K}_t = \cup_{i=0}^t \mathcal{C}_i $ and given a sequence of $S$ incremental steps, 
our goal is to learn a model 
mapping input images to either their corresponding label in $\mathcal{K}_S$ or to the special class $unk$. 
In the following we will split the classification model into two components: a feature extractor $f$ that maps the samples into a feature space and a classifier $g$ that maps the features into a class label, i.e. $g(f(x))=c$ with $c \in \{\mathcal{K}_S,unk\}$.


\subsection{Preliminaries}
Standard approaches to tackle the OWR problem apply non-parametric classification algorithms on top of learned metric spaces \cite{bendale2015towards, de2016online}. A common choice for the classifier $g$ is the Nearest Class Mean (NCM) \cite{mensink2012metric, guerriero2018deep}. 
NCM works by computing a centroid for each class (i.e. the mean feature vector) and assigning a test sample to the closest centroid in the learned metric space. Formally, we have:
\begin{equation}
    g^{\text{NCM}}(x)=\argmin_{c \in \mathcal{C}_t} d(f(x), \mu_c)
\end{equation}
where $d(\cdot,\cdot)$ is a distance function (e.g. Euclidean) and $\mu_c$ is the mean feature vector for class $c$. The standard NCM formulation cannot be applied in the OWR setting since it lacks the inherent capability of detecting images belonging to unknown categories. To this extent, in \cite{bendale2015towards} the authors extend the NCM algorithm to the OWR setting by defining a rejection criterion for the unknowns. In this extension, called Nearest Non-Outlier (NNO), class scores are defined as: 
\begin{equation} 
\label{eq:nno-score} s_c^\mathtt{NNO}(x)=\mathcal{Z} (1-\frac{d(f(x), \mu_c)}{\tau}), 
\end{equation}
where $\tau$ is the rejection threshold and $\mathcal{Z}$ is a normalization factor. The final classification is held-out as:
\begin{equation}
\label{eq:rejection-nno}
    g(x) = 
  \begin{cases}
    unk & \text{if}\; s_c^\mathtt{NNO}(x) \leq 0\; \forall{c \in \mathcal{K}_t},\\
    g^{\text{NCM}}(x) & \text{otherwise}.
  \end{cases}
\end{equation}
{Following \cite{mensink2012metric}, in \cite{bendale2015towards} the features are linearly projected into a metric space defined by a matrix $W$ (i.e. $f(x)=W\cdot x$)}, 
with $W$ learned on the first training set $\mathcal{T}_0$ and kept fixed during the successive learning steps.
The main limitation of this approach is that new knowledge will be incorporated in the classifier $g$ without updating the feature extractor $f$ accordingly. 
In \cite{mancini2019knowledge}, it is shown how the performance of NNO can be significantly improved by using as $f$ a deep architecture trained end-to-end in each incremental step.
The proposed algorithm, DeepNNO, trains the deep neural network by minimizing the binary cross-entropy loss: 
\begin{equation} 
\label{eq:deep-nno-prob}
    \ell(x_i,c_i)= \sum_{c\in\mathcal{C}_t} \mathds{1}_{c=c_i} \log(s_c^\mathtt{DNNO}(x_i))+ \mathds{1}_{c\neq c_i} \log(1-s_c^\mathtt{DNNO}(x_i))
\end{equation} 
where $s_c^\mathtt{DNNO}(x)$ is the class scores computed as $s_c^\mathtt{DNNO}(x)=e^{-\frac{1}{2} ||f(x) - \mu_c||^2}$.
Differently from \cite{mensink2012metric,bendale2015towards}, the underlying feature representation of the data changes along with the parameters of the backbone architecture. As a consequence, it is not possible to fix the class-specific centroids, especially in the incremental learning setting, since changes in the network parameters will create a shift among the computed old class centroids and the current network activations. Such shift  cannot be recovered, since the training sets $\mathcal{T}_i$ with $i<t$ are not available. 
To overcome this problem, DeepNNO proposes to (i) update online the class centroids and (ii) perform rehearsal using as memory stored samples of old classes. 
Additionally, DeepNNO uses the network at the previous learning step to compute a distillation loss \cite{hinton2015distilling, rebuffi2017icarl} on the network activations, reducing the catastrophic forgetting problem by preventing them from deviating from the features used to discern old classes. 

Finally, \cite{mancini2019knowledge} updates online the rejection threshold during training with an heuristic rule that raises the threshold whenever the network predicts true positives or negatives and lowers it whenever the network predicts false positives or negatives. The final classification is held-out as in Eq.\eqref{eq:rejection-nno}.

While we will base our architecture and classifier on \cite{mancini2019knowledge}
, we argue that DeepNNO has two main drawbacks. First, the learned feature representation $f$ is not forced to produce predictions clearly \textit{localized} in a limited region of the metric space. 
{Indeed, constraining the feature representations of a given class to a limited region of the metric space allows to have both more confident predictions on seen classes and producing clearer rejections also for images of unseen concepts.} Second, having an heuristic strategy for setting the threshold is sub-optimal with no guarantees on the robustness of the choice. In the following, we will detail how we provide solutions to both problems.

 \begin{figure}[t]
    \centering
    \centering
    \includegraphics[width=\linewidth]{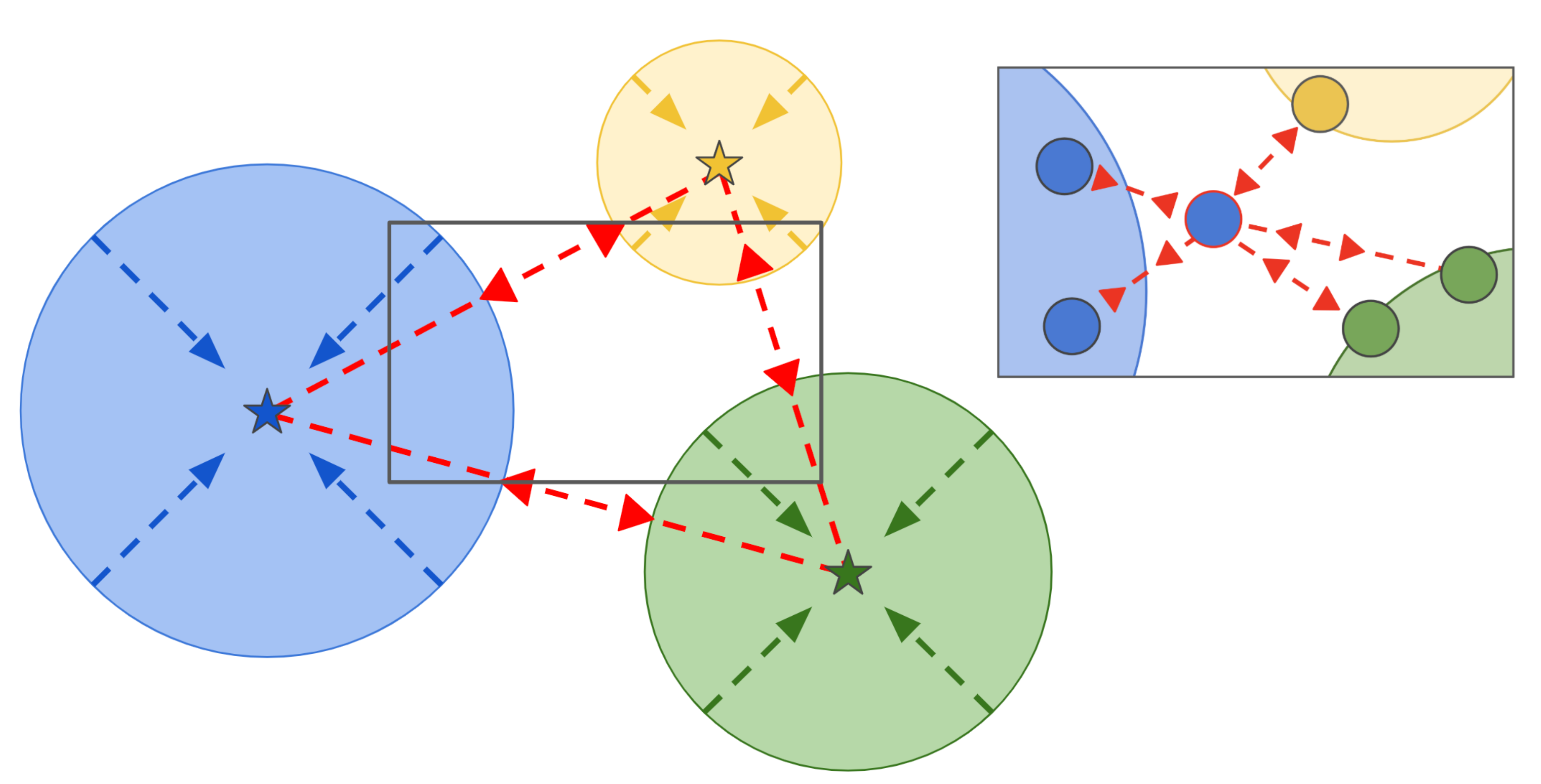} 
    \hfill
    \caption{Overview of the proposed global to local clustering. The global clustering (left) pushes sample representations closer to the centroid (star) of the class they belong to. 
    The local clustering (right), instead, forces the neighborhood of a sample in the representation space to be semantically consistent, pushing away samples of other classes. \vspace{-15pt}}
    \label{fig:method1}
\end{figure}

\subsection{Boosting Deep Open World Recognition}
{To obtain feature representations clearly localized in the metric space based on their semantic, 
we propose to use a pair of losses enforcing clustering.} 
In particular, we use a \textit{global} term which forces the network to map samples of the same class close to their centroid (Fig.\ref{fig:method1}, left) and a \textit{local} clustering term which constrains the neighborhood of a sample to be semantically consistent, i.e. to contain samples of the same class (Fig.\ref{fig:method1}, right). 
In the following we describe the two clustering terms. 

\myparagraph{Global Clustering}.
The global clustering term aims to reduce the distance between the features of a sample with the centroid of its class. To model this, we took inspiration from what has been proposed in \cite{mensink2012metric} and we employ a cross-entropy loss with the probabilities obtained through the distances among samples and class centroids.
Formally, given a sample $x$ and its class label $c$, we define the global clustering term as follows:
\begin{equation} 
    \ell_{GC}(x, c) = - \log \frac{s_c(x)}{ \displaystyle\sum_{k \in \mathcal{K}_t} s_k(x)}.
\end{equation}
The class-specific score $s_c(x)$ is defined as:
\begin{equation} \label{eq:prob-softmax} 
    s_c(x) = \frac{e^{-\frac{1}{T}||f(x) - \mu_c||^2}}{ \displaystyle\sum_{k \in \mathcal{C}_t}  e^{- \frac{1}{T}||f(x) - \mu_k||^2}}
\end{equation}
where $T$ is a temperature value which allows us to control the behavior of the classifier. We set $T$ as the variance of the activations in the feature space, $\sigma^2$, in order to normalize the representation space and increase the stability of the system. 
During training, $\sigma^2$ is the variance of the features extracted from the current batch while, at the same time, we keep an online global estimate of $\sigma^2$ that we use at test time. 
The class mean vectors $\mu_i$ with $i \in \mathcal{K}_t$ as well as $\sigma^2$ are computed in an online fashion, as in \cite{mancini2019knowledge}. 

\myparagraph{Local Clustering}. 
To enforce that the neighborhood of a sample in the feature space is semantically consistent (i.e. given a sample $x$ of a class $c$, the nearest neighbours of $f(x)$ belong to $c$), we employ the soft nearest neighbour loss \cite{salakhutdinov2007learning, frosst2019analyzing}. 
{This loss has been proposed to measure the class-conditional entanglement of features in the representation space}. 
In particular, it has been defined as: 
\begin{equation}
    \label{SNNL}
    \ell_{LC}(x, c,\mathcal{B})  = - \log\ \  
    \frac{ \displaystyle\sum_{\mathclap{x_j \in \mathcal{B}_c\setminus \{x\}}
                               }
                      e^{- \frac{1}{T}||f(x) - f(x_j)||^2}}{
          \displaystyle\sum_{\mathclap{x_k \in \mathcal{B}\setminus \{x\}}}
             e^{- \frac{1}{T}||f(x) - f(x_k)||^2}
          }
\end{equation}
where T refers to the temperature value, $\mathcal{B}$ is the current training batch, and $\mathcal{B}_c$ is the set of samples in the training batch belonging to class $c$. Instead of performing multiple learning steps to optimize the value of T as proposed in \cite{frosst2019analyzing}, we use as $T=\sigma^2$ as we do in Eq.~\ref{eq:prob-softmax}. 

Intuitively, given a sample $x$ of a class $c$, a low value of the loss indicates that the nearest neighbours of $f(x)$ belong to $c$, while high values indicates the opposite (i.e. nearest neighbours belong to classes $i\in\mathcal{K}_t$ with $i\neq c$). 
Minimizing this objective allows to enforce the semantic consistency in the neighborhood of a sample in the feature space.

\myparagraph{Reducing catastrophic forgetting through distillation.}
As highlighted in the previous sections, to avoid forgetting old knowledge, we want the feature extractor to preserve the behaviour learned in previous learning steps. 
To this extent, we follow standard rehearsal-based approaches for incremental learning \cite{rebuffi2017icarl, chaudhry2018riemannian, mancini2019knowledge, castro2018end} and we introduce (i) a memory which stores the most relevant samples for classes in $\mathcal{K}_t$ and (ii) a distillation loss which enforces consistency among the features extracted by $f$ and ones obtained by the feature extractor of the previous learning step, $f_{t-1}$. 
Formally, the distillation loss is computed as:
\begin{equation}
    \ell_{DS}(x) =||f(x)- f_{t-1}(x)||.
\end{equation}
This loss is minimized only for incremental training steps, hence, only when $t>1$. 

Overall, we train the network to minimize on a batch of samples $\mathcal{B}={\{(x_1 , c_1), \cdots, (x_{|\mathcal{B}|}, c_{|\mathcal{B}|})}\}$ the following loss:
\begin{equation} \label{eq:final_loss}
    \mathcal{L}=\frac{1}{|\mathcal{B}|} \sum_{(x,c) \in \mathcal{B}} \ell_{GC}(x,c) + \lambda\ \ell_{LC}(x,c,\mathcal{B}) + \gamma\ \ell_{DS}(x)
\end{equation}
with $\lambda$ and $\gamma$ hyperparameters weighting the different components. We set $\lambda=\gamma=1$ in all experiments. 

\begin{figure}[t]
    \centering
    \includegraphics[width=0.7\linewidth]{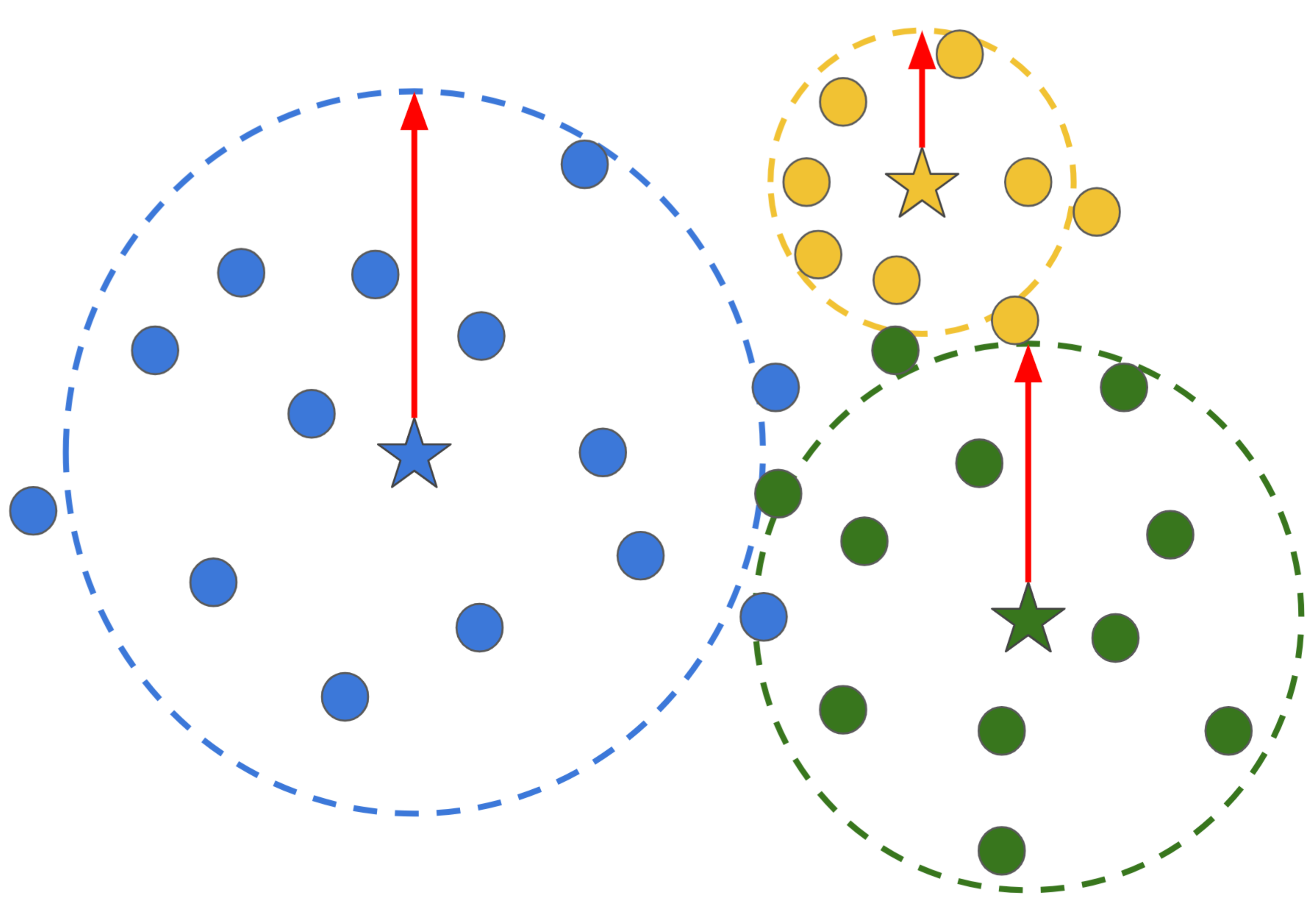} 
    \caption{
    Overview of the learning of the \textit{class-specific} rejection thresholds. The small circles represent the samples in the held out set. The dashed circles, having radius the maximal distance (red), represent the limits beyond which a sample is rejected as a member of that class. As it can be seen, the class-specific threshold is learned to reduce the rejection errors. Best viewed in colors. \vspace{-15pt}} 
    \label{fig:method2}
\end{figure}

\myparagraph{Learning to detect the unknown}.
In order to extend our NCM-based classifier to work on the open set scenario, we explicitly learn class-specific rejection criterions. As illustrated in Fig.~\ref{fig:method2}, for each class $c$ we define the \textit{class-specific} threshold as the maximal distance $\Delta_c$ for which the sample belongs to $c$. Under this definition, our classifier is:
\begin{equation}
\label{eq:rejection-our}
    g(x) = 
  \begin{cases}
    unk & \text{if}\, d(f(x),\mu_c)> \Delta_c,\; \forall{c \in \mathcal{K}_t},\\
    \text{argmin}_c d(f(x),\mu_c)&\text{otherwise}
  \end{cases}
\end{equation}
with $d(x,y)=\frac{1}{\sigma^2}||x-y||^2$. 
Instead of heuristically estimating or fixing a maximal distance, we explicitly learn it for each class minimizing the following objective:
\begin{equation}
    \ell_{MD}(x,c) = \sum_{k \in \mathcal{K}_t} \max (0, m \cdot (\frac{1}{\sigma^2}||f(x) - \mu_k||^2 - \Delta_k))
\end{equation}
where $m=-1$ if $c=k$ and $m=1$ otherwise. 
The $\ell_{MD}$ loss leads to an increase of $\Delta_c$ if the distance from a sample belonging to the class $c$ and the class centroid $\mu_c$ is greater than $\Delta_c$. Instead, if a sample not belonging to $c$ has a distance from $\mu_c$ less then $\Delta_c$, it increases the value of $\Delta_c$.

{Overall, the training procedure of our method is made of two steps: in the first we train the feature extractor on the training set minimizing Eq.~\ref{eq:final_loss}, while in the second we learn the distances $\Delta_c$ on a set of samples which we held-out from training set.
To this extent, we split the samples of the memory in two parts, one used for updating the feature extractor $f$ and the centroids $\mu_c$ 
and the other part for learning the $\Delta_c$ values.}

\section{Experiments} 
In this section, we first introduce the experimental setting and the metrics used for the evaluation, then we report results of our experiments and an ablation study of our contributions.

 \begin{figure*}[t]
    \vspace{5pt}
    \centering
    \begin{subfigure}{0.245\linewidth}
        \includegraphics[width=\linewidth]{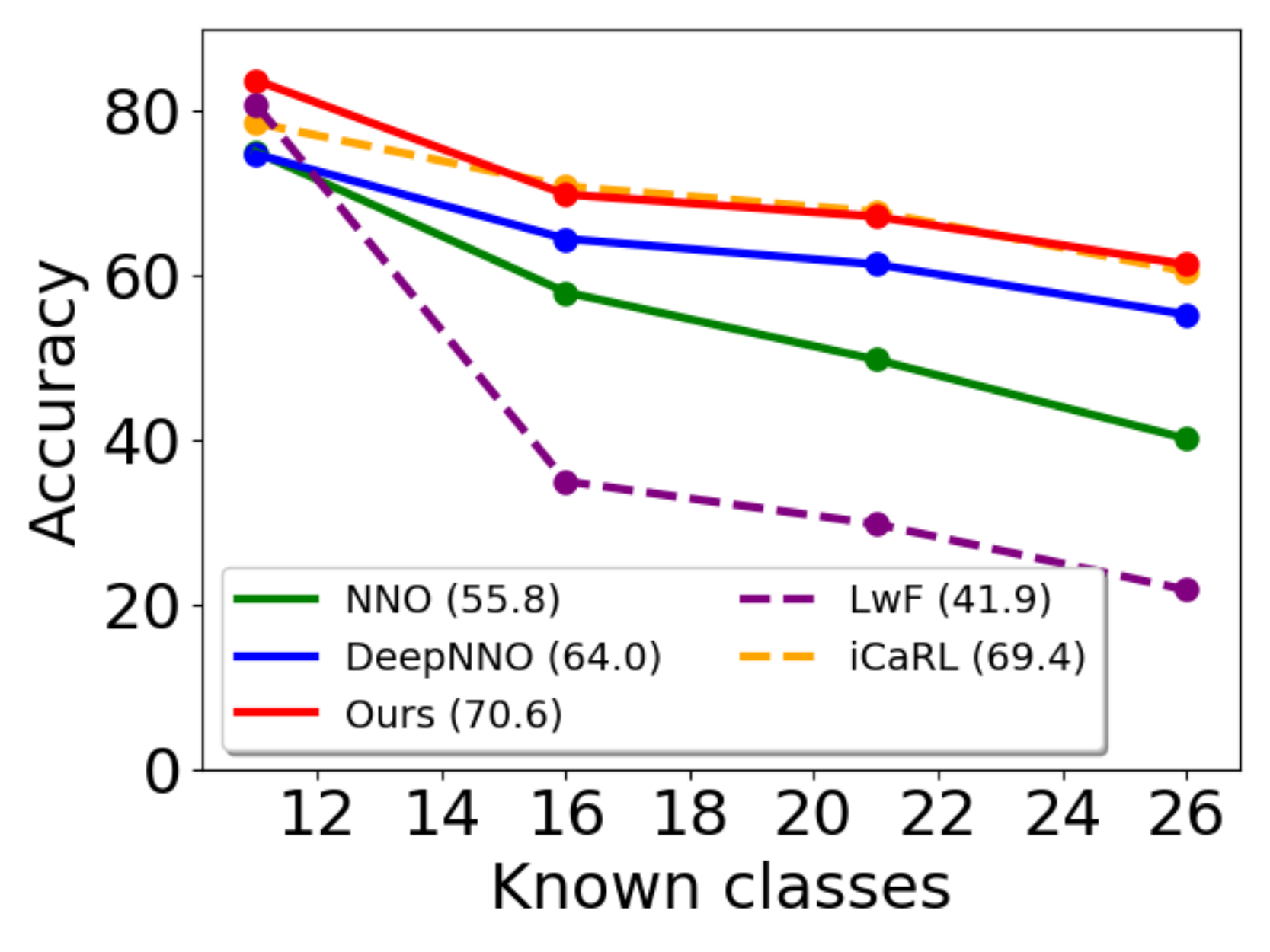}
        \caption{\centering{Closed World Without Rejection}\vspace{-5pt}}
        \label{fig:ROD-wor}
    \end{subfigure}
    \hfill
    \begin{subfigure}{0.245\linewidth}
        \includegraphics[width=\linewidth]{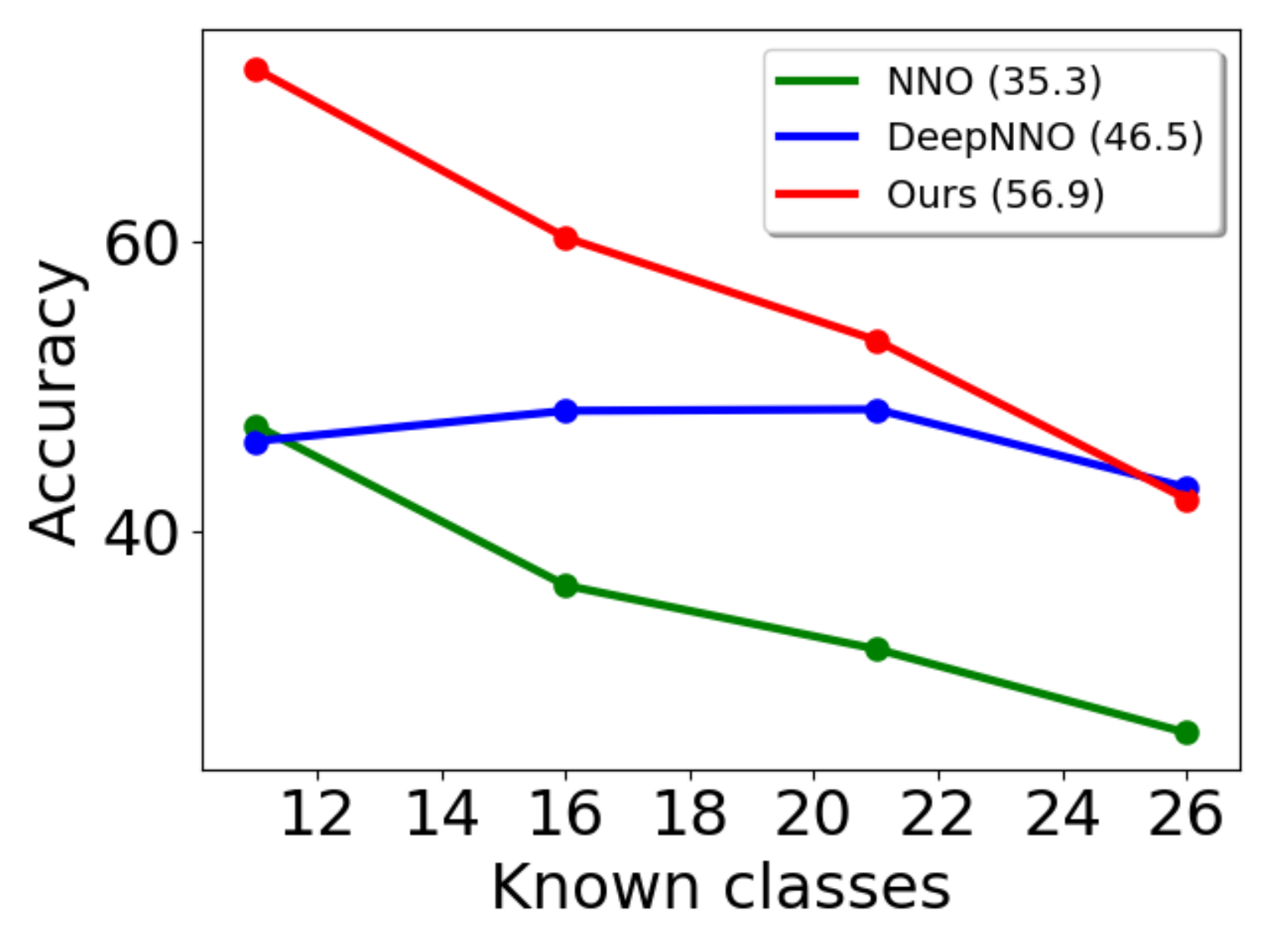}
        \caption{\centering{Closed World With Rejection}\vspace{1em}\vspace{-5pt}}
        \label{fig:ROD-wir}
    \end{subfigure}
    \hfill
    \begin{subfigure}{0.245\linewidth}
        \includegraphics[width=\linewidth]{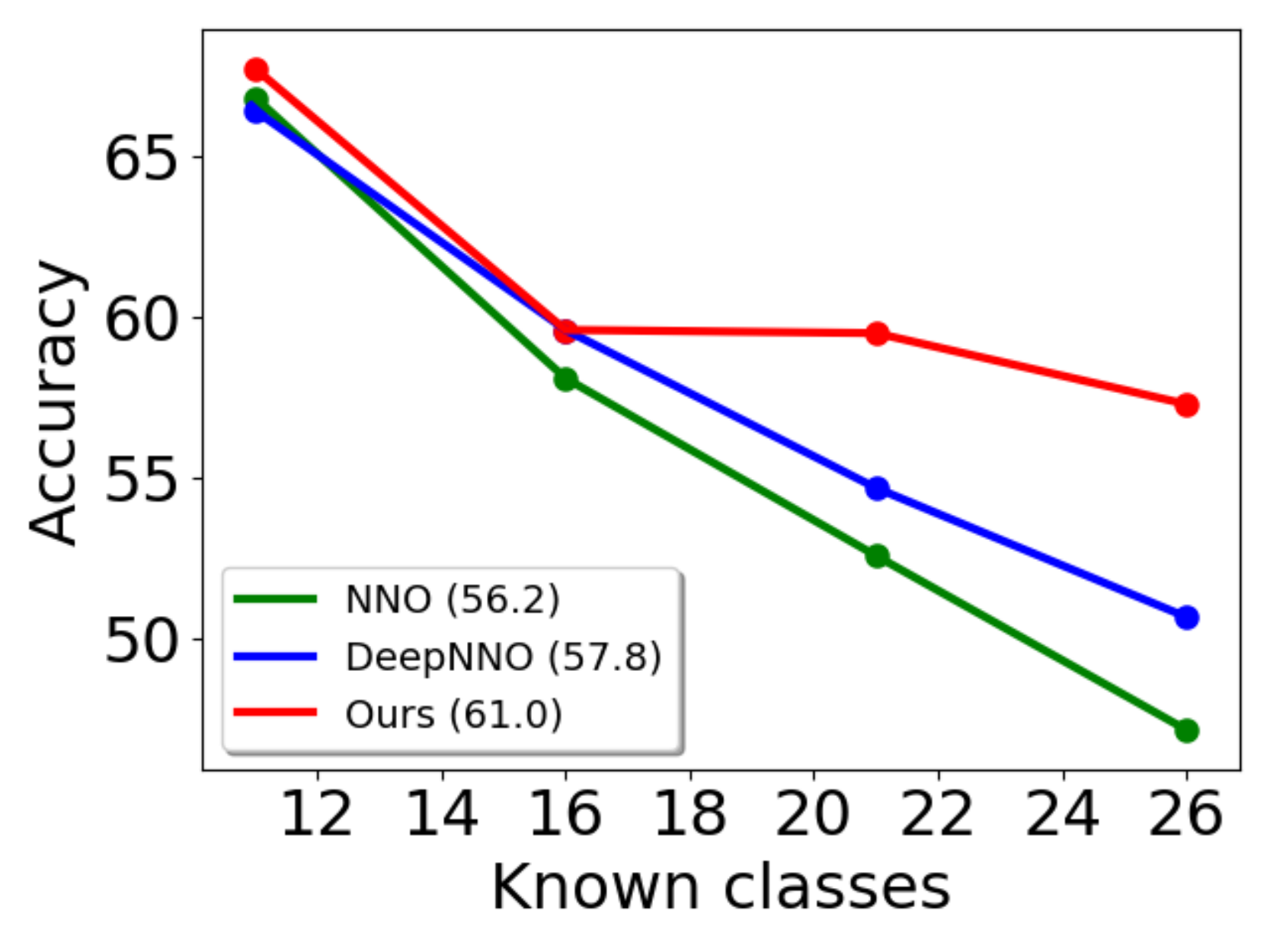}
        \caption{\centering{Open World Recognition Average}\vspace{-5pt}}
        \label{fig:ROD-owra}
    \end{subfigure}
   \hfill
    \begin{subfigure}{0.245\linewidth}
        \includegraphics[width=\linewidth]{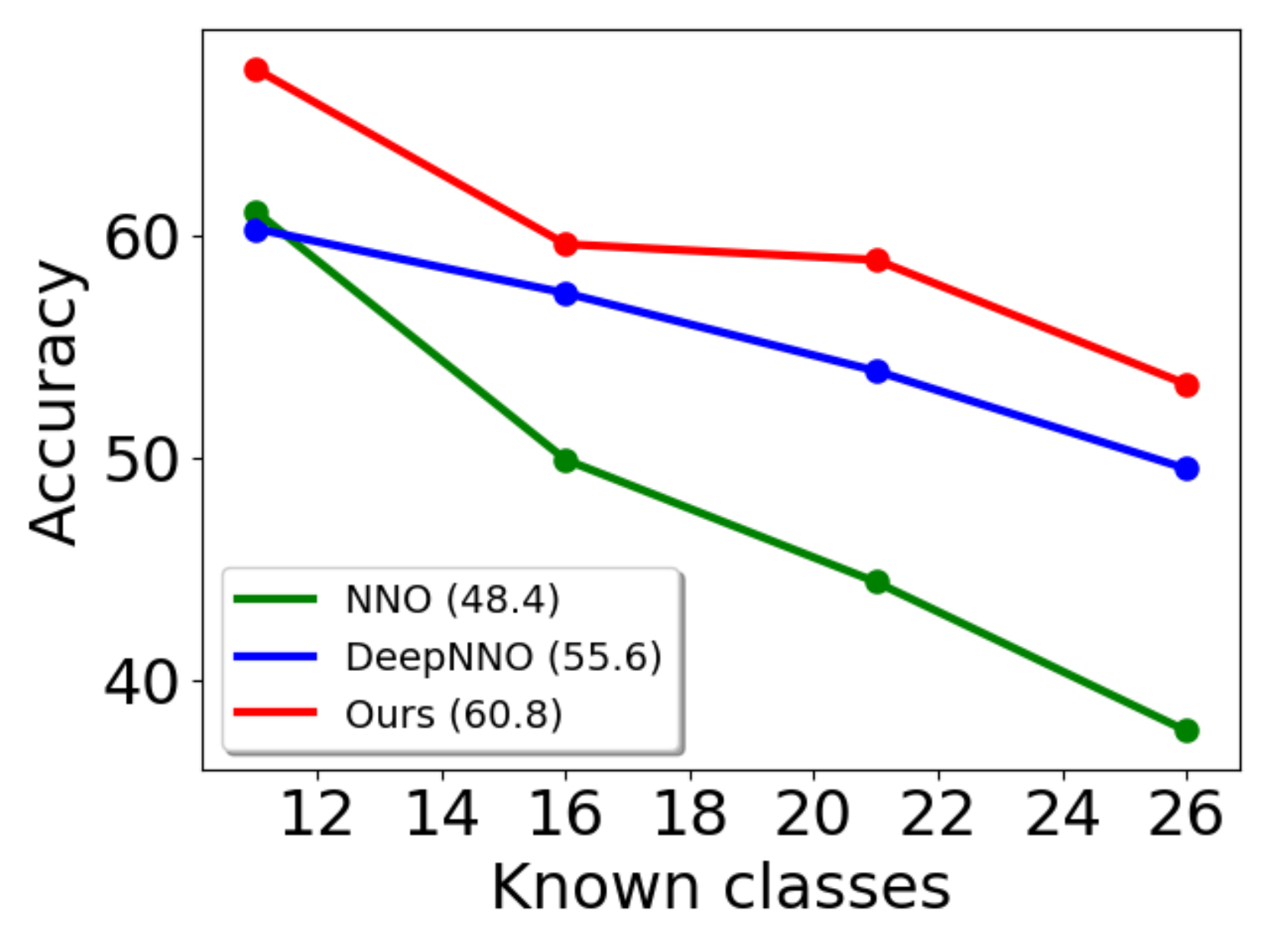}        
        \caption{\centering{Open World Recognition Harmonic Mean}\vspace{-5pt}}
        \label{fig:ROD-owrh}
    \end{subfigure}
     \caption{Comparison of NNO \cite{bendale2015towards}, DeepNNO \cite{mancini2019knowledge} and our method on RGB-D Object dataset \cite{lai2011large}. The numbers in parenthesis denote the average accuracy among the different incremental steps. \vspace{-13pt}}
     \label{fig:ROD}
\end{figure*}

 \begin{figure*}[t]
    \centering
    \begin{subfigure}{0.245\linewidth}
        \centering
        \includegraphics[width=\linewidth]{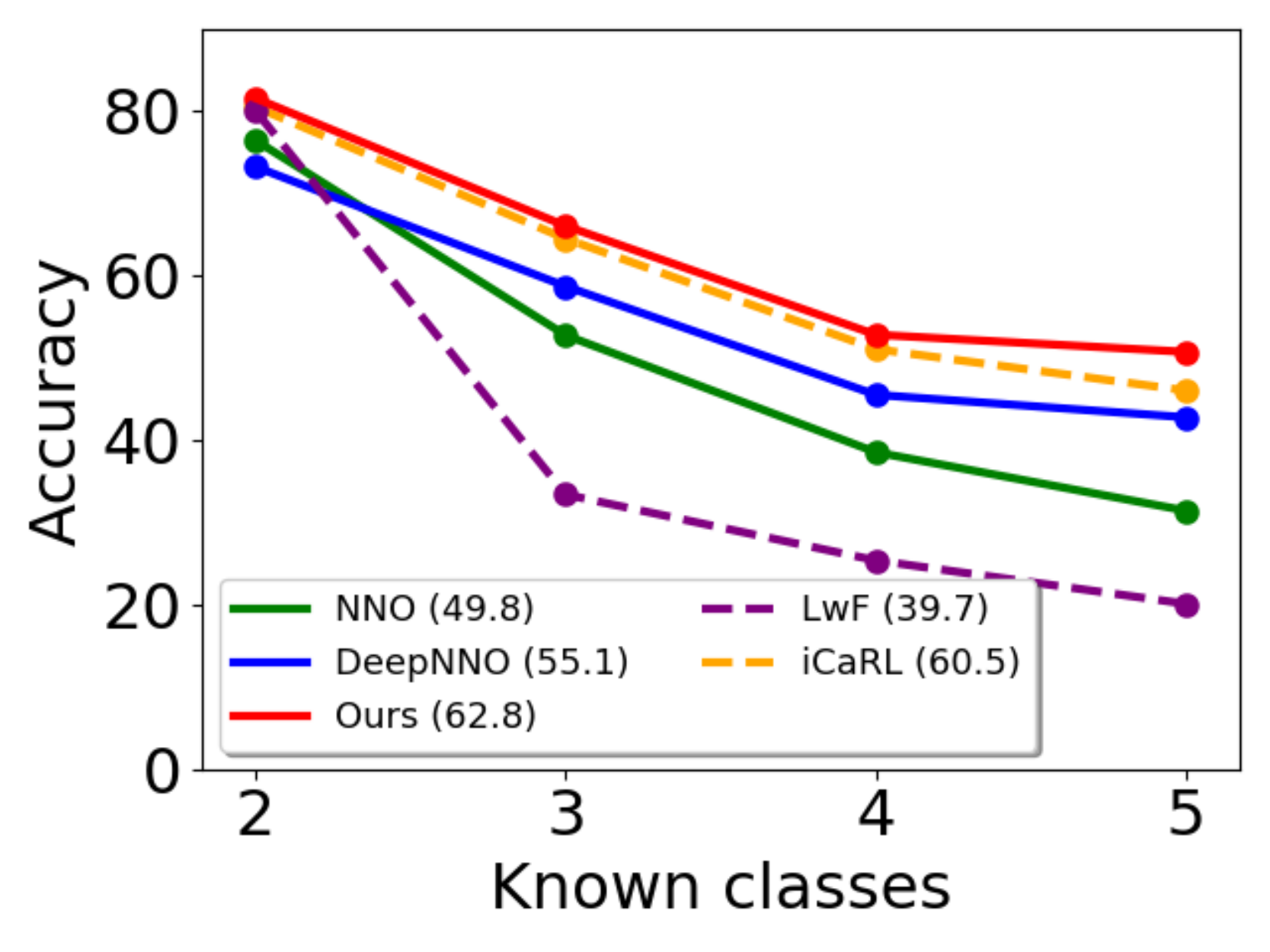}
        \caption{\centering{Closed World Without Rejection}\vspace{-5pt}}
        \label{fig:Core-wor}
    \end{subfigure}
    \hfill
    \begin{subfigure}{0.245\linewidth}
        \centering
        \includegraphics[width=\linewidth]{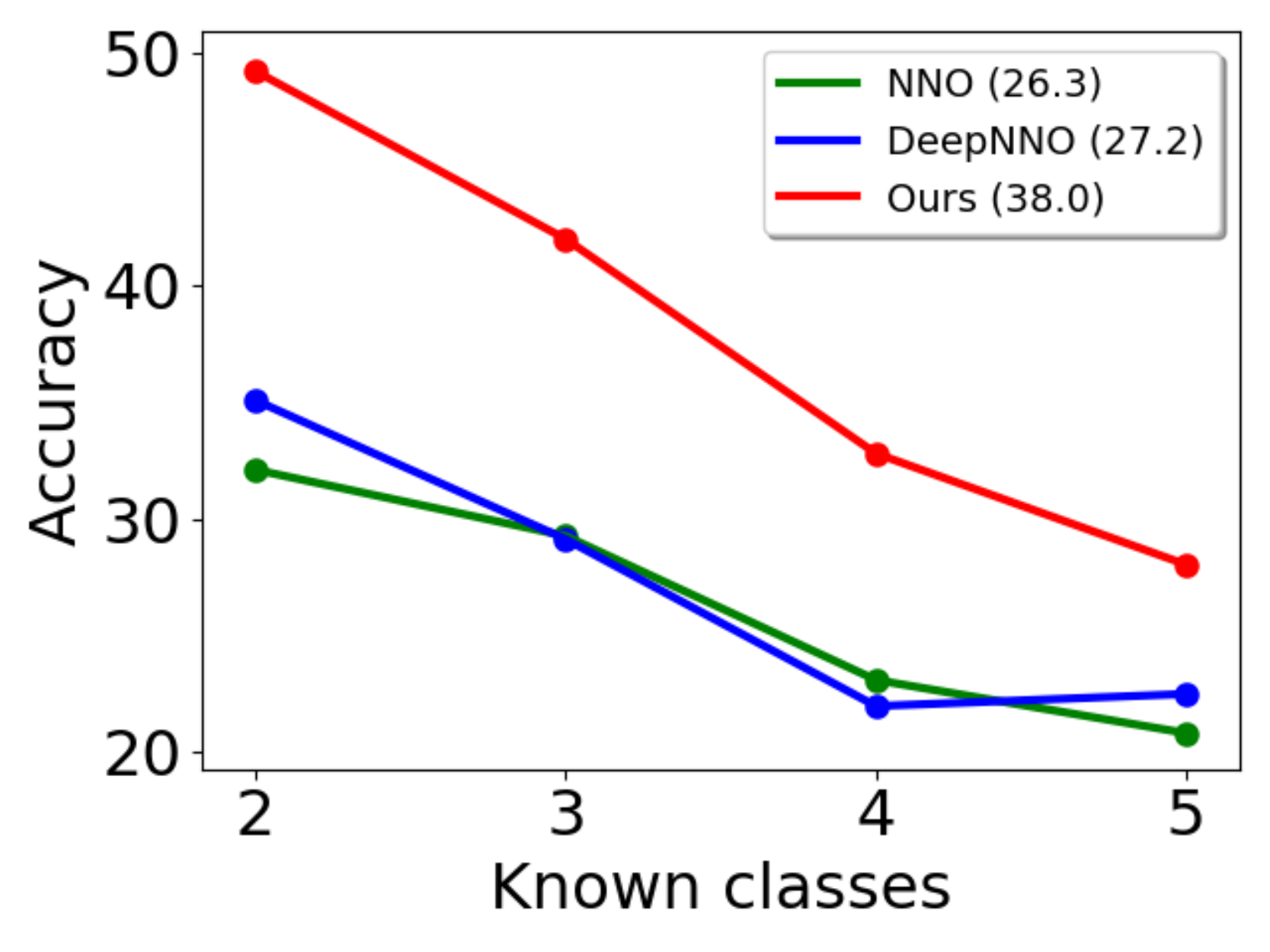}
        \caption{\centering{Closed World With Rejection}\vspace{1em}\vspace{-5pt}}
        \label{fig:Core-wir}    
    \end{subfigure}
    \hfill
    \begin{subfigure}{0.245\linewidth}
        \centering
        \includegraphics[width=\linewidth]{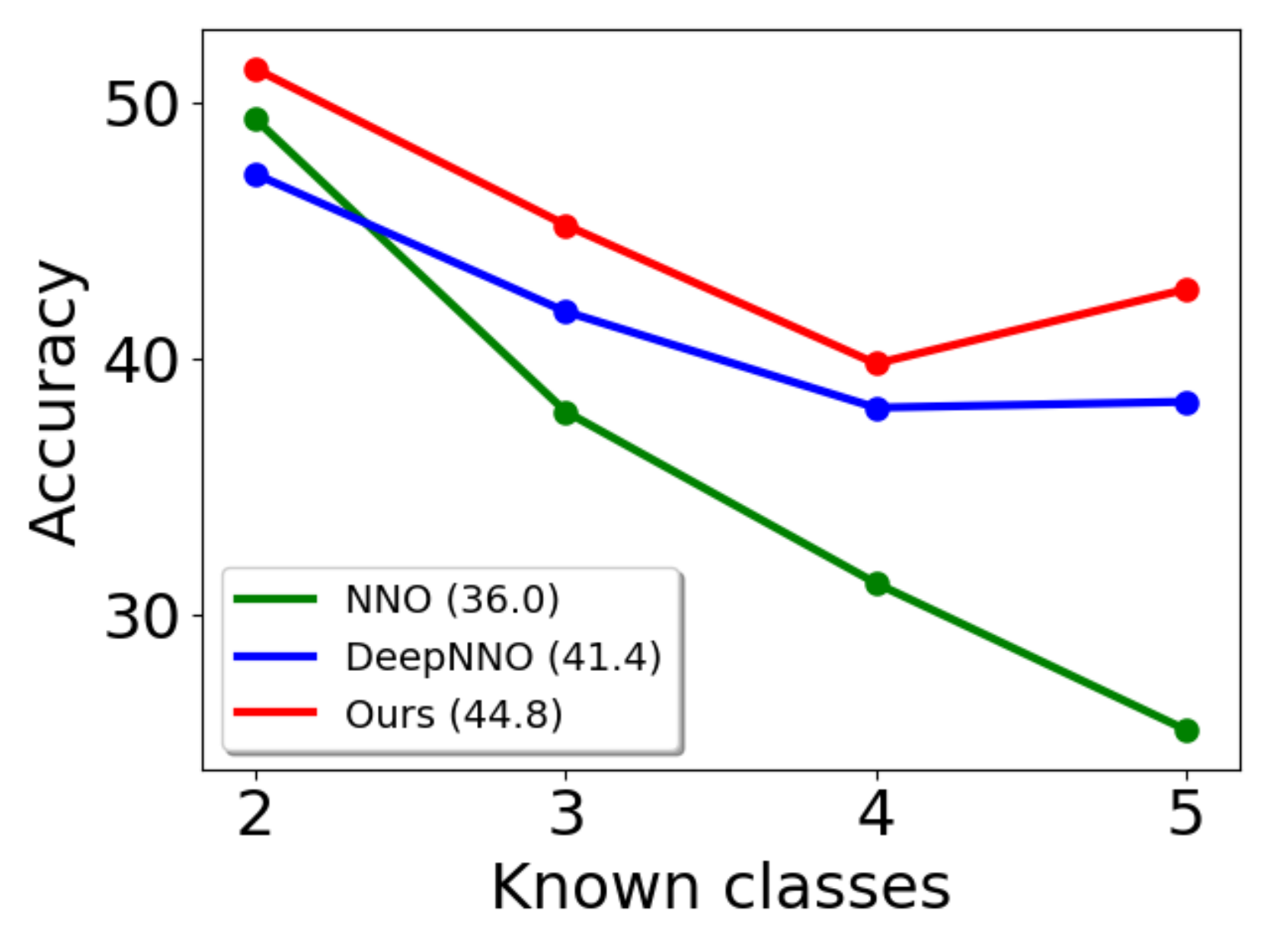}
        \caption{\centering{Open World Recognition Average}\vspace{-5pt}}
        \label{fig:Core-owra}    
        \end{subfigure}
    \hfill
    \begin{subfigure}{0.245\linewidth}
        \centering
        \includegraphics[width=\linewidth]{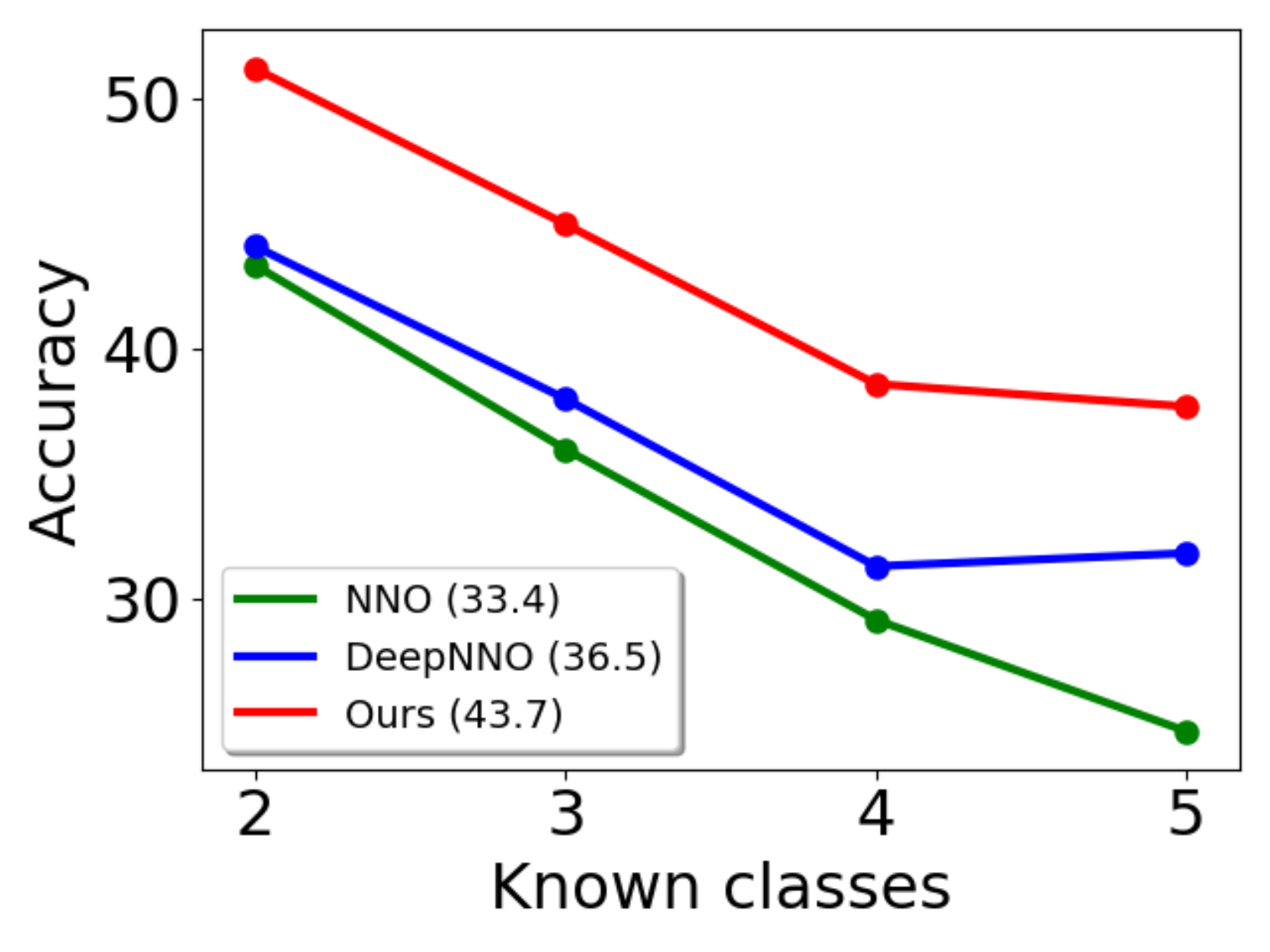}        
        \caption{\centering{Open World Recognition Harmonic Mean}\vspace{-5pt}}
        \label{fig:Core-owrh}
    \end{subfigure}
    \caption{Comparison of NNO \cite{bendale2015towards}, DeepNNO \cite{mancini2019knowledge} and our method on Core50 dataset \cite{lomonaco2017core50}. The numbers in parenthesis denote the average accuracy among the different incremental steps. \vspace{-13pt}}
    \label{fig:Core}
\end{figure*}

\begin{figure}[t]
    \centering
    \begin{subfigure}{0.49\linewidth}
        \centering
        \includegraphics[width=\linewidth]{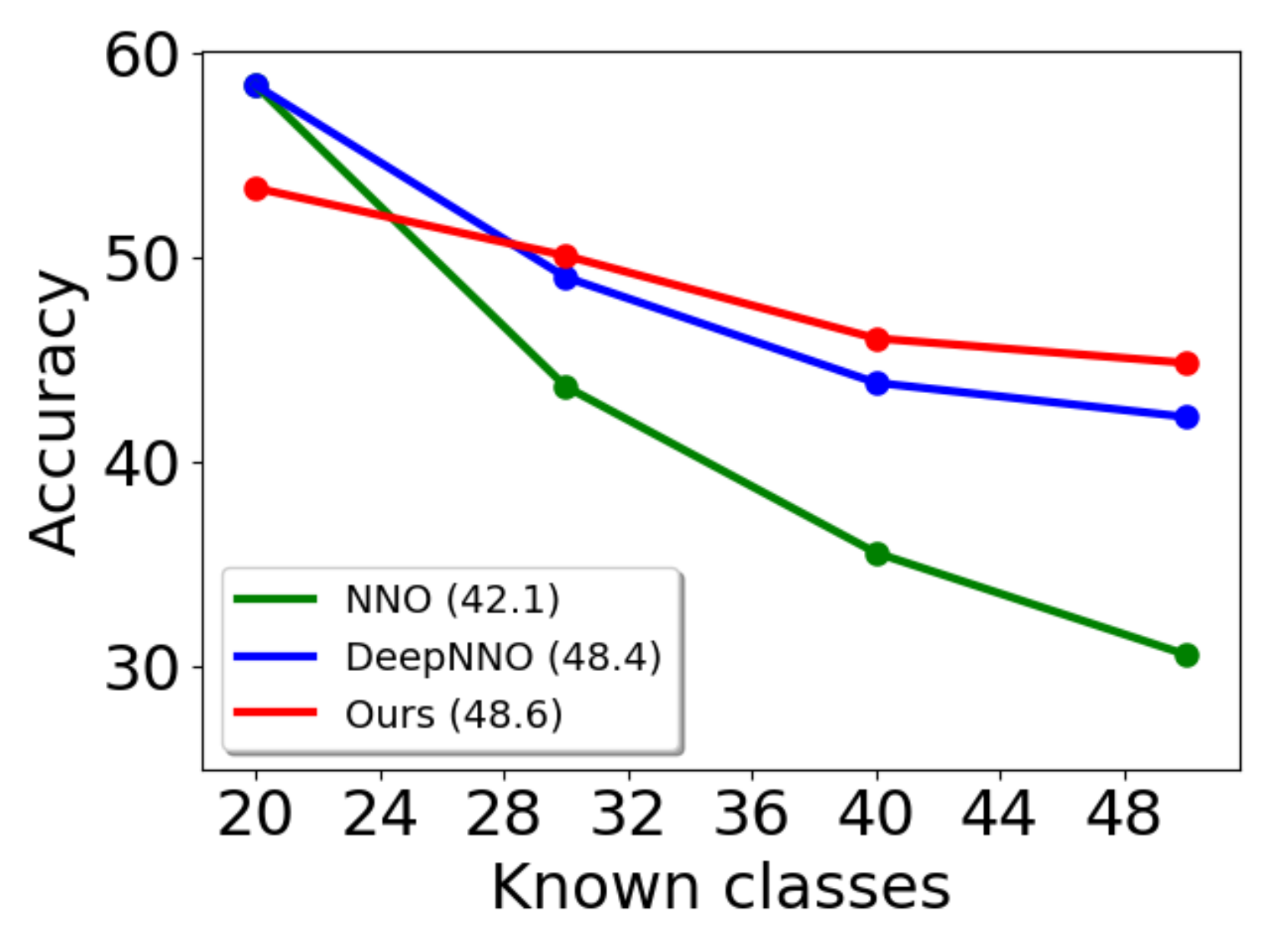}
        \caption{\centering{Open World Recognition Average}\vspace{-5pt}}
        \label{fig:cifar-owra}    
        \end{subfigure}
    \hfill
    \begin{subfigure}{0.49\linewidth}
        \centering
        \includegraphics[width=\linewidth]{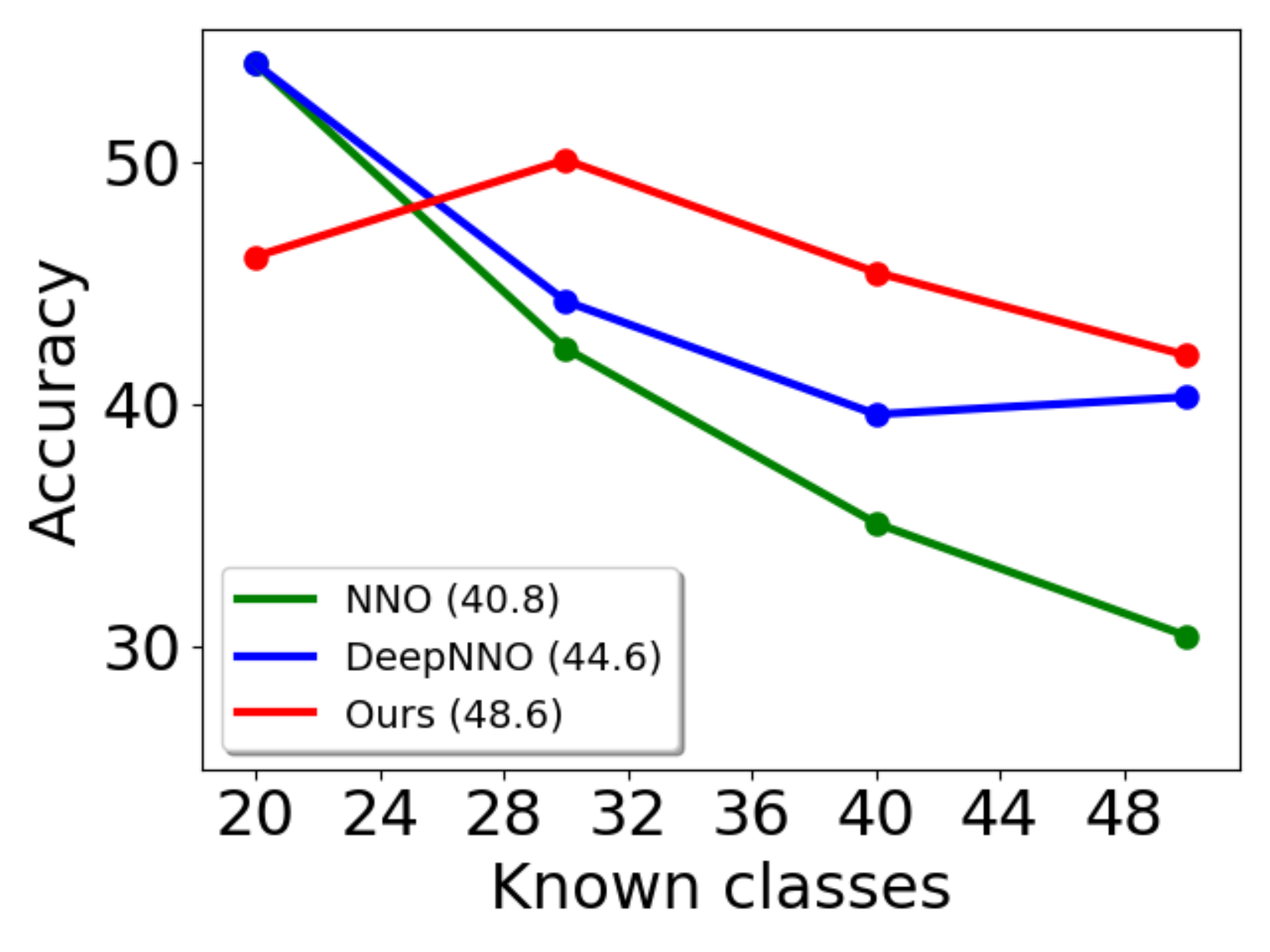}        
        \caption{\centering{Open World Recognition Harmonic Mean}\vspace{-5pt}}
        \label{fig:cifar-owrh}
    \end{subfigure}
    \caption{Comparison of NNO \cite{bendale2015towards}, DeepNNO \cite{mancini2019knowledge} and our method on Cifar dataset \cite{krizhevsky2009learning}. The numbers in parenthesis denote the average accuracy among the different steps. \vspace{-13pt}}
    \label{fig:cifar}
\end{figure}

\subsection{Experimental Setting}

\myparagraph{Datasets and Baselines.} We assess the performance of our model on three datasets: RGB-D Object \cite{lai2011large} Core50 \cite{lomonaco2017core50} and CIFAR-100 \cite{krizhevsky2009learning}. The RGB-D Object dataset \cite{lai2011large} is one of the most used dataset to evaluate the ability of a model to recognize daily-life objects. It contains 51 different semantic categories that we split in two parts in our experiments: 26 classes are considered as known categories, while the other 25 are the set of unknown classes. Among the 26 classes, we consider the first 11 classes as the initial training set and we incrementally add the remaining classes in 4 steps of 5 class each. 
{As proposed in \cite{lai2011large}, we sub-sample the dataset taking one every fifth frame. For the experiments, we use the first train-test split among the original ones defined by the authors \cite{lai2011large}. In each split one object instance from each class is chosen to be used in the test set and removed from the training set.} This split provides nearly 35,000 training images and 7,000 test images.
Core50 \cite{lomonaco2017core50} is a recently introduced benchmark for testing continual learning methods in an egocentric setting. The dataset contains images of 50 objects grouped into 10 semantic categories. The images have been acquired on 11 different sequences with varying conditions. 
Following the standard protocol described in \cite{lomonaco2017core50}, we select the sequences 3, 7, 10 for the evaluation phase and use the remaining ones to train the model. Due to these differences in conditions between the sequences, Core50 represents a very challenging benchmark for object recognition. As in the RGB-D Object dataset, we split it into two parts: 5 classes are considered known and the other 5 as unknown. In the known set, the first 2 classes are considered as the initial training set. The others are incrementally added 1 class at a time. 
CIFAR-100 \cite{krizhevsky2009learning} is a standard benchmark for comparing
incremental class learning algorithms \cite{rebuffi2017icarl}. It contains 100 different semantic categories. We follow previous works \cite{mancini2019knowledge} splitting the dataset into 50 known and 50 unknown classes and considering 20 classes as the initial training set. Then, we incrementally add the remaining ones in steps of 10 classes. We evaluate the performance of our method in the OWR scenario comparing it to DeepNNO \cite{mancini2019knowledge} and NNO \cite{bendale2015towards}, using the implementation in \cite{mancini2019knowledge} for the latter. We further compare our method with two standard incremental class learning algorithms, namely LwF \cite{li2017learning} (in the MC variant of \cite{rebuffi2017icarl}) and iCaRL \cite{rebuffi2017icarl}. Both LwF and iCaRL are designed for the closed world scenario, thus we use their performances as reference in that setting, without open-ended evaluation. 
For each dataset, we have randomly chosen five different sets of known and unknown classes. After fixing them, we run the experiments three times for each method. The results are obtained by averaging the results among each run and order.

\myparagraph{Networks architectures and training protocols.} Following previous works, we use a ResNet-18 architecture \cite{he2016deep} for all the experiments. For RGB-D Object dataset and Core50, we train the network from scratch on the initial classes for 12 epochs and for 4 epochs in the incremental steps. For CIFAR-100, instead, we set the epochs to 120 for the initial learning stage and to 40 for each incremental step.
We use a learning rate of $0.1$ and batch size 128 for the RGB-D Object dataset, while we use $0.01$ and 64 for Core50. We train the network using Stochastic Gradient Descent (SGD) with momentum $0.9$ and a weight decay of $10^{-3}$ on both datasets.We resize the images of RGB-D Object dataset to $64 \times 64$ pixels and the images of Core50 to $128 \times 128$ pixels. We perform random cropping and mirroring for all the datasets. Moreover, for the set of held-out samples, we also perform color jittering varying brightness, hue and saturation.
For the baselines, we use the same network architecture and training protocol defined in \cite{mancini2019knowledge}.
We also employ the same strategy for memory management, considering a fixed size of 2000 samples and constructing each batch by drawing 40\% of the instances from memory. Differently from \cite{mancini2019knowledge}, we never see during training 20\% of the samples present in memory, using them only to learn the values the class-specific threshold values $\Delta_k$.

\myparagraph{Metrics}
We use 3 standard metrics for comparing the performances of OWR methods. For the closed world we show the global accuracy \textit{with} and \textit{without} rejection option. Specifically, in the closed world \textit{without rejection} setting, the model is tested only on the known set of classes, \textit{excluding} the possibility to classify a sample as \textit{unknown}. This scenario measures the ability of the model to correctly classify samples among the given set of classes. In the closed world \textit{with rejection} scenario, instead, the model can either classify a sample among the known set of classes or categorize it as \textit{unknown}. This scenario is more challenging than the previous one because samples belonging to the set of known classes might be misclassified as \textit{unknowns}. For the open world we use the standard OWR metric defined in \cite{bendale2015towards} as the average between the accuracy computed on the closed world \textit{with rejection} scenario and the accuracy computed on the open set scenario (i.e. the accuracy on rejecting samples of unknown classes). Since the latter metric creates biases on the final score (i.e. a method rejecting every sample will achieve a 50\% accuracy), we introduced the OWR-H as the harmonic mean between the accuracy on open set and the closed world with rejection scenarios to mitigate this bias.
\subsection{Quantitative results}
We report the results on the RGB-D Object dataset in Fig.~\ref{fig:ROD}. 
Considering 
the closed world without rejection, reported in Fig.~\ref{fig:ROD-wor}, we note that our method is able to improve the feature representation, outperforming DeepNNO by 5.6\% of accuracy on average and NNO by 14.8\%. The reason for the improvement comes from the introduction of the global and local clustering loss terms, which allows the model to better aggregate samples of the same class and to better separate them from samples of other classes. Comparing our model with the incremental class learning approaches LwF and iCaRL, we can see that our approach is highly competitive, surpassing LwF with a large gap while being comparable with the more effective iCaRL. We believe these are remarkable results given that the main goal of our model is not to purely extend its knowledge over time with new concepts. The comparison on the closed world with rejection, shown in Fig.~\ref{fig:ROD-wir}, demonstrates that our method is also more confident on the known classes, being able to reject a lower number of known samples. In particular, our method is more confident on the first incremental steps, and obtains, on average, an accuracy of 10.3\% more than DeepNNO.
Considering the open world metrics, our method is superior to previous works. From the results of OWR, reported in Fig.~\ref{fig:ROD-owra}, we see that our method reaches performance similar to DeepNNO in the first steps, while it outperforms it in the latest ones.
However, considering the OWR-H (Fig.~\ref{fig:ROD-owrh}), our method is better in all the incremental steps. This is because 
previous methods are biased towards rejecting more samples, as it is demonstrated by the lower closed world with rejection performance they achieve. On the contrary, our learned rejection criterion, coupled with our clustering losses, allows to achieve a better trade-off between the accuracy of open set and closed world with rejection. 
Overall, our method improves 
on average by 4.8\% and 5.2\% with respect to DeepNNO in the OWR and OWR-H metrics respectively.

In Fig.~\ref{fig:Core} we report the results on the Core50 \cite{lomonaco2017core50} dataset. Similarly to the RGB-D Object dataset, our method achieves competitive results with respect to incremental class learning algorithms designed for the closed world scenario, remarkably outperforming iCaRL by 4.7\% of accuracy in the last incremental step. It also achieves a superior performance in both closed world, without and with rejection option with respect to state-of-the-art OWR algorithms, outperforming NNO by 13.01\% and DeepNNO by 7.74\% on average in the first (Fig.~\ref{fig:Core-wor}) and by more than $10\%$ for both NNO and DeepNNO in the latter (Fig.~\ref{fig:Core-wir}).
In particular, it is worth noting how both DeepNNO and NNO are not able to properly model the confidence threshold, rejecting most of the sample of the known classes. Indeed, by including the rejection option the accuracy drops to 27.2\% and 26.3\% respectively for DeepNNO and NNO, while our model reaches an average accuracy of 38.0\%. 
In Fig.~\ref{fig:Core-owra} and Fig.~\ref{fig:Core-owrh}, we report the OWR performances (standard and harmonic) on Core50. {Our method outperforms DeepNNO by 3.4\% and 7.2\% in average respectively in standard OWR and OWR-H metrics}, confirming the effectiveness of our clustering losses and learned class-specific maximal distances. 

Finally, in Fig. \ref{fig:cifar} we report the results on the CIFAR-100 dataset in terms of the OWR (Fig. \ref{fig:cifar-owra}) and OWR-H metrics (Fig. \ref{fig:cifar-owrh}). Even in this benchmark, our approach achieves superior results, on average, than previous methods. Our model achieves lower performances with respect to NNO and DeepNNO only in the initial training stage. 
However, 
in the incremental learning steps our model clearly outperforms both methods, demonstrating its ability to learning and recognizing in an open-world without forgetting old classes. In fact, considering the incremental steps, the average improvement of our model over NNO are of ~10\% in both OWR and OWR-H metrics, while over DeepNNO are of 2\% for the OWR and 4.5\% for the OWR-H metric.

\subsection{Ablation study.} Our approach is mainly built on three components, i.e. global clustering loss (GC), local clustering loss (LC) and the learned class-specific {rejection thresholds}.
In this section we analyze each proposed contribution. We start from the two clustering losses and then we compare the choice we made for the rejection with other common choices.


\begin{table}[]
\vspace{5pt}
\centering
\begin{tabular}{l||llll||l|l}
\multicolumn{1}{c||}{\textbf{Method}} & \multicolumn{4}{c||}{\textbf{Known Classes}} & \multicolumn{2}{c}{\textbf{OWR}} \\
 & \multicolumn{1}{c}{11} & \multicolumn{1}{c}{16} & \multicolumn{1}{c}{21} & \multicolumn{1}{c||}{26} & \multicolumn{1}{c}{{[}20{]}} & \multicolumn{1}{c}{H} \\\hline 
\multicolumn{1}{l||}{GC} & \multicolumn{1}{l|}{66.0} & \multicolumn{1}{l}{57.3} & \multicolumn{1}{l|}{58.6} & 53.3 & 58.8 & \multicolumn{1}{l}{58.7} \\ 
\multicolumn{1}{l||}{LC} & \multicolumn{1}{l|}{64.1} & \multicolumn{1}{l}{56.0} & \multicolumn{1}{l|}{57.9} & 56.4 & 58.6 & \multicolumn{1}{l}{58.4} \\\multicolumn{1}{l||}{Triplet} & \multicolumn{1}{l|}{62.1} & \multicolumn{1}{l}{54.9} & \multicolumn{1}{l|}{54.8} & 49.5 & 55.4 & \multicolumn{1}{l}{55.4} \\  
\multicolumn{1}{l||}{GC + LC} & \multicolumn{1}{l|}{\textbf{67.7}} & \multicolumn{1}{l}{\textbf{59.6}} & \multicolumn{1}{l|}{\textbf{59.5}} & \textbf{57.3} & \textbf{61.0} & \multicolumn{1}{l}{\textbf{60.8}} \\ 
\end{tabular}
\vspace{-5pt}
\caption{Ablation study on the global (GC), local clustering (LC) and {Triplet} loss on the OWR metric. The right column shows the average OWR-H over all steps. 
}
\label{tab:ablation_clustering}
\vspace{-5pt}
\end{table}


\begin{table}[]
\begin{tabular}{l|cc|ll|l}
\multirow{2}{*}{\textbf{Method}} &\textbf{Class} &\textbf{Multi} & \multirow{2}{*}{\textbf{Known}} & \multirow{2}{*}{\textbf{Unknown}} & \multirow{2}{*}{\textbf{Diff.}} \\ 
&\textbf{specific} & \textbf{stage}& & & \\\hline
 {DeepNNO [}22{]}& & & 84.4 & 98.8 & 14.4 \\\hline
\multirow{3}{*}{Ours} & \cmark & & 83.0 & 98.6 & 15.6 \\
& &\cmark & 4.4 & 26.9 & 22.6 \\
 & \cmark& \cmark& 27.4 & 65.2 & \textbf{37.8} \\  \hline
\end{tabular}
\vspace{-5pt}
\caption{Rejection rates of different techniques for detecting the unknowns. The results are computed using the same feature extractor on the RGB-D Object dataset.
}
\label{tab:ablation_unknown}
\vspace{-15pt}
\end{table}

\myparagraph{Global and local clustering.} In Table \ref{tab:ablation_clustering} we compare the two clustering terms considering the open world recognition metrics in the RGB-D Object dataset. By analyzing the two loss terms separately we see that, on average, they show similar performance. In particular, using only the global clustering (GC) term we achieve slightly better performance on the first three incremental steps, while on the fourth the local clustering (LC) term is better.
However, the best performance on every step is achieved by combining the global and local clustering terms (GC + LC). This demonstrates that the two losses provide different contributions, being complementary to learn a representation space which properly clusters samples of the same classes while better detecting unknowns. Finally, since our loss functions and triplet loss \cite{balntas2016learning} share the same objective, i.e. building a metric space where samples sharing the same semantic are closer then ones with different semantics, we report in Table \ref{tab:ablation_clustering} also the results achieved by replacing our loss terms with a triplet loss \cite{balntas2016learning}. 
As the Table shows, the triplet loss formulation (Triplet) fails in reaching competitive results with respect to our full objective function, with a gap of more than 5\% in both standard OWR metric and OWR harmonic mean. Notably, it achieves lower results also with respect to all of the loss terms in isolation and the superior performances of LC confirm the advantages of SNNL-based loss functions with respect to triplets, as shown in \cite{frosst2019analyzing}.

\myparagraph{Detecting the Unknowns.} In Table \ref{tab:ablation_unknown} we report a comparison of different strategies to reject samples on the RGB-D Object dataset \cite{lai2011large}. In particular, using the same feature extractor, we compare the proposed method to learn the class-specific maximal distances with three baselines: (i) we adopt the strategy proposed by DeepNNO \cite{mancini2019knowledge}, (ii) we learn class-specific maximal distances but during training (i.e. without our two-stage pipeline) and (iii) we learn a single maximal distance which applies to all classes using our two-stage training strategy. 
The comparison is performed considering the difference of the rejection rates on the known and unknown samples. For the known class samples, we report the percentage of correctly classified samples in the closed-world that are rejected when the rejection option is included.
We intentionally remove the wrongly classified samples since we want to isolate rejection mistakes from classification ones. On the unknown samples, we report the open-set accuracy, i.e. the percentage of rejected samples among all the unknown ones. In the third column, we report the difference among the open-set accuracy and the rejection rate on known samples. Ideally, the difference should be as close as possible to 100\%, since we want a 100\% rejection rate on unknown class samples and 0\% on the known class ones.
From the table, we see that the highest gap is achieved by the class-specific maximal distance with the two-stage pipeline we proposed, which rejects 27.4\% of known class samples and 65.2\% on the unknown ones. The gap with the other strategies is remarkable. Using the {two stage-pipeline but a class-generic maximal distance} leads to a low rejection rate, both on known and unknown samples, achieving a difference of 22.6\%, which is 15.2\% less than using a class-specific distance. On the other hand, estimating the confidence threshold as proposed in DeepNNO \cite{mancini2019knowledge} or without our two-stage pipeline provides a very high rejection rate, both on known and unknown classes, which lead to a difference of 14.4\% and 15.6\% for DeepNNO and the single-stage strategy respectively,  the lowest two among the four strategies. 
In fact, computing the thresholds using only the training set biases the rejection criterion on the overconfidence that the method has acquired on this set. At test time, this causes the model to consider the different data distribution (caused by e.g. different object instances) as a source for rejection even if the actual concept present in the input is known. Using the two-stage process allows to overcome this bias, tuning the rejection criterion on unseen data on which the model cannot be overconfident.

\section{Conclusion and Future Works} 
In this work we presented an approach to tackle the open world recognition problem in robot vision. As in previous works, we base our approach on a NCM classifier built on top of deep features, and  we boost the OWR performances of this framework by training the deep architecture to minimize 
 a global to local semantic clustering loss. This loss allows to reduce distances of samples of the same class in the feature space while separating them from points belonging to other classes, thus better detecting unknown concepts. Moreover, we avoid heuristic estimates of a rejection criterion for detecting unknowns by explicitly learning class-specific distances beyond which a sample is rejected. Quantitative and qualitative analysis on standard recognition benchmarks show the efficacy of our approach and choices, outperforming previous state-of-the-art OWR algorithms. 
While here we considered the OWR scenario, there are still many directions that could be explored for enabling robots to learn autonomously in the real word. One could be extending the approach to the Web-aided OWR scenario considered in \cite{mancini2019knowledge}. In particular, when training images are autonomously retrieved from the Web, they come with an inherent noisy labelling. In this context our model will not be able to work well since the noise will be transmitted to the class-specific cluster center hampering the efficacy of the recognition model. 
 Moreover, it would be interesting to analyze the OWR problem in an active learning context \cite{parisi2019rethinking}. 
 Finally, another interesting research direction would be extending OWR approaches to more complex tasks, such as object detection and segmentation, where the ability 
 to distinguish among background/stuffs \cite{lin2014microsoft} and actual unknown objects and the background shift problem illustrated in \cite{cermelli2020modeling}. 

\bibliography{bib}

\end{document}